# Robotic self-assessment of competence

Gertjan J. Burghouts
Intelligent Imaging
TNO
The Netherlands
gertjan.burghouts@tno.nl

Albert Huizing
Hybrid AI
TNO
The Netherlands
albert.huizing@tno.nl

Mark A. Neerincx
TNO-PCS / TU-Delft Interactive Intelligence
The Netherlands
mark.neerincx@tno.nl

## ABSTRACT

In robotics, one of the main challenges is that the on-board Artificial Intelligence (AI) must deal with different or unexpected environments. Such AI agents may be incompetent there, while the underlying model itself may not be aware of this (e.g., deep learning models are often overly confident). This paper proposes two methods for the online assessment of the competence of the AI model, respectively for situations when nothing is known about competence beforehand, and when there is prior knowledge about competence (in semantic form). The proposed method assesses whether the current environment is known. If not, it asks a human for feedback about its competence. If it knows the environment, it assesses its competence by generalizing from earlier experience. Results on real data show the merit of competence assessment for a robot moving through various environments in which it sometimes is competent and at other times it is not competent. We discuss the role of the human in robot's self-assessment of its competence, and the challenges to acquire complementary information from the human that reinforces the assessments.

## 1 Introduction

In robotics, one of the main challenges is that the on-board Artificial Intelligence (AI) gets confronted with various environments while moving through the world. Such AI agents may be incompetent in environments that are different from the environments it was optimized for during the training of the AI model (especially for deep-learning based techniques [1]), and these models may produce very unexpected predictions in unknown environments, due to the domain gap [2] or the out-of-distribution problem [3]. Moreover, deep learning models typically are often overconfident [4]. This is especially problematic for (semi-) autonomous systems, because they have a degree of autonomy and need to be reliable.

It is crucial for the moving AI agent that it has a capability to assess whether the agent knows the current environment, and, if so, if it is competent there. We refer to this capability as 'competence assessment' (CA). Competence relates to the agent's specific task at hand. The interpretation of what competence means for this task, needs to come from an external source (human or external system). In this work, we consider CA based on camera images and world knowledge.

Section 2 motivates the need for CA. Section 3 describes a method for CA when nothing is known beforehand, whereas Section 4 describes CA when prior knowledge about competence is available. Results are shown in Section 5. Section 6 discusses the role of CA in human-robot interaction. Section 7 concludes the paper.

## 2 Competence Assessment

If the agent does not know the current environment, then it also cannot know its competence there. Hence it needs to take action in order to get to know the environment and its competence therein. This can be achieved by asking a human for feedback about its current competence. With this feedback, it updates the model of environments and associated competences.

If the agent knows the current environment and it knows that it is competent here, then it can safely continue its operation. On the contrary, if the agent knows that it is incompetent, then it can resolve to actions such as asking for assistance by a human, or, moving back to an environment where it knows that it is competent or safe (e.g., home, or the previous environment). The assistance of the human can be to help the agent to navigate, to assist with its task, or to provide supervision to improve the underlying AI model such that it becomes competent here.

Below we describe the CA module, i.e., estimating the probability that the AI model knows the current environment, and, if so, if it is competent there. Two variants of CA have been developed: (1) When nothing is known about (in)competences beforehand 'CA-zero' (section 3), and (2) when there is expert knowledge about (in)competences of the AI model, 'CA-expert' (section 4). In the current implementation of the robot car, it is assumed that nothing is known beforehand, hence CA-zero is applied. Further, it is assumed that, in the case of an unknown environment, the human is asked for feedback about the competence.

## 3 CA-zero: when nothing is known about (in)competences beforehand

The rationale of CA-zero is that it assesses competence by relating the current environment to previously observed environments and its competence in those environments. To that end, CA-zero measures how similar the current environment is to previously observed environments. When the AI agent starts, it has not observed anything, so the human needs to provide feedback about competence. When the AI agent starts to move, the environment changes, and CA-zero measures the similarity to observed environments (subsection 3.1). The similarity is calibrated (subsection 3.2) and transformed, in order to arrive at a probability that the environment is known, P(known) (subsection 3.3). CA-zero checks if P(known) is sufficiently large, depending on a pre-set threshold, to decide if it knows the current environment. If it knows the environment, CA-zero also assesses the probability that it is competent, based on prior knowledge about competence in previously observed environments (subsection 3.3). If CA-zero does not know the environment, it asks the human for feedback about competence. That feedback is used to update CA-zero's model of environments and competence (subsection 3.4). In the next subsections, we detail each of the reasoning steps.

### 3.1 Similarities

The current environment is transformed into a description, i.e., a feature vector. This transformation is performed by a convolutional neural network (CNN). Since it has to run on a moving platform with limited resources, we have adopted MobileNetV2 [5], a light-weighted CNN with good accuracy. To compare with previously observed environments, the L2 norm is considered as a pair-wise distance between two vectors of respective environments. This L2 norm is calibrated and then fed to a Gaussian probability distribution function [6] to obtain a probability measure of how well an environment is known (subsection 3.2). This probability measure is used to assess the current environment. When the current environment is known, the competence is assessed (subsection 3.3). When the current environment is not known, feedback is required, and CA-zero's model is updated (subsection 3.4).

### 3.2 Calibration

To have a sense of similarity distances between environments and to convert them to sensible probabilities, they are calibrated. This calibration is conducted by looking at distances between many environments sampled from the MIT Places database [7]. The average nearest neighbor is often related, e.g., a park and a forest. Therefore, the average nearest neighbor is mapped to a probability of P(known) = 0.5 using a Gaussian function:

P( known | E, M, S ) = exp(-argmin{F from M\E}( D(E, F) )^2 / S^2)

with E the description vector of an environment in the MIT Places database, M all the other description vectors of the MIT Places database, D the Euclidean distance between two description vectors, F the nearest neighbor of E, and S approximated by S* to ensure that on average P( known | E ) = 0.5 for all E and their respective nearest neighbors F in the MIT Places database.

This mapping function is calibrated during the initialization prior to the robot's operation. Figure 1 shows that this function is effective to assess P(known). In the upper left, there are different environments from 'buildings', such as office spaces, effectively mapping them to P(known) ~ 0.

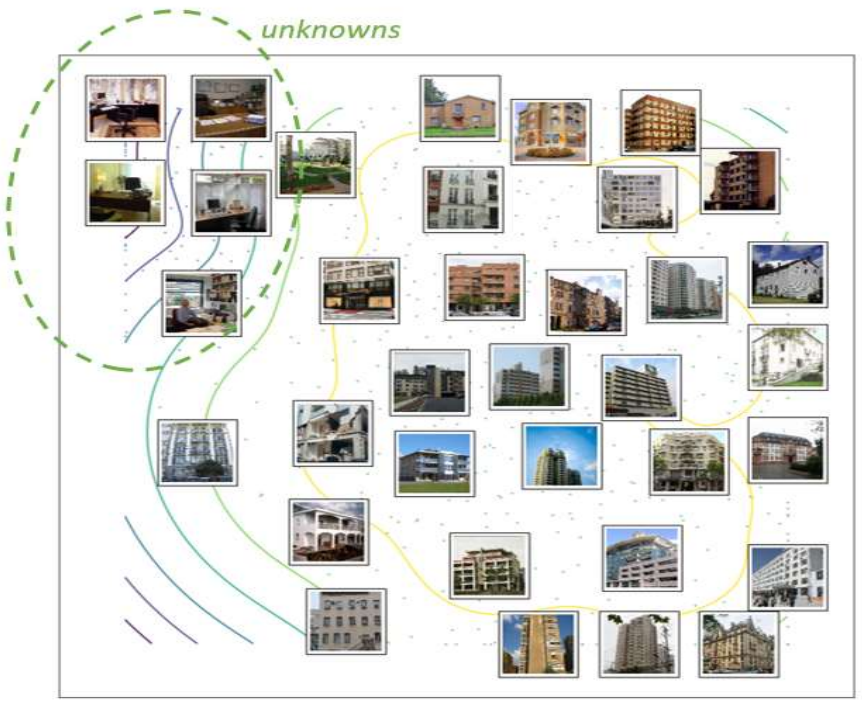

**Figure 1: Visual relations and P(known) is high for related environments (buildings, middle) and low for unrelated environments (e.g., office, upper left).**

### 3.3 P(known) and P(competent)

A new environment G is transformed into a feature vector, and the equation P and parameter S* from subsection 3.2 are used to estimate P( known | G, M, S* ),

P( known | G, H, S* ) = exp(-argmin{F from M}( D(G, F) )^2 / S*^2)

with G the description vector of the current environment, and H description vectors of earlier seen environments. When the robot starts, it has not seen any environments, hence: H = {} and P = 0. In this case, it does not know the environment (P = 0), hence it cannot assess its competence. It asks the human for feedback about its competence. This also happens when H ≠ {} but all environments are different from the current environment: P ≈ 0. In that case, it also asks the human for feedback.

On the contrary, if P ≈ 1, the environment is known. This means that it can assess its competence, P(competent). P(competent) is based on P(known). If the nearest neighbor has the connotation of 'competent' (based on the feedback of the human for that environment), then the P(competent) is mapped to [0, 1] (where 1 means competent), otherwise to [0, -1] (where -1 means incompetent), such that the robot can make a distinction between competence and incompetence on a continuous scale.

## 3.4 Adaptability

Recapitulating, if the current environment is not known, the agent asks the human for feedback about its competence in that environment. After feedback, CA-zero's model is updated, by adding the description vector and the competence into CA-zero's database, so the environment is known at a later stage. The probability function of subsection 3.2 ensures generalization due to its spread (parameter S*). Also, the description vector generalizes to some extent, because it was optimized to be similar for the same type of environments. These two properties enable the agent to generalize the competence assessment to similar environments.

While the robot is moving, it may observe environments that are known, but after a while it may enter a different, unknown environment. After feedback about competence, all intermediate observed environments are associated to either endpoint and its associated competence. In this way, the model is refined to also include information about competence in slowly varying environments.

# 4 CA-expert: when (in)competence is known for specific environments

The rationale of CA-expert is that it assesses competence by relating the current environment to world knowledge. This world knowledge consists of a set of statements about environments and associated competence. For example, 'is not competent in nature environments'.

Note that this approach is very different from CA-zero, which assumes no prior knowledge and can only relate environments visually and with very low-level feedback about competence for a particular environment. On the contrary, CA-expert can reason about high-level statements about competence and these can be included prior to the robot's operation.

CA-expert relates the current environment to the statements about known competence, in order to infer the competence. This relation is based on both appearance (visual) and the textual description (semantic). Suppose, that the current environment looks like a park (visual). A park is similar to the concept of 'nature' (semantic). CA-expert knows that the AI model is not competent in nature environments. Combining the visual and semantic relatedness gives an indication of the competence, P(competent). This combination is implemented by reasoning in a visual-semantic embedding (subsection 4.1). To be able to compare visual to semantic information, a normalization needs to be done (subsection 4.2), in order to arrive at P(competent) (subsection 4.3).

If the current environment does not relate to any of the statements, CA-expert needs to get external knowledge, such as asking the human for feedback. This is similar to CA-zero. The advantage of CA-expert (compared to CA-zero) is that the AI agent starts with some knowledge of competence. In the next subsections, we detail each of the reasoning steps.

## 4.1 Visual-Semantic Embedding

CA-expert relates the current environment to the statements about known competence in particular environments. To that end, it needs to reason about environments. We do this through a semantic space, a word2vec embedding [8]. Figure 2 shows that similar environments are effectively grouped. The environment 'nature' (red) relates to forest, mountain, sky, etc. (lower right) and not to campus, podium, car, etc. (middle). Related environments are closer, which enables to reason about environments based on their distance in this embedding.

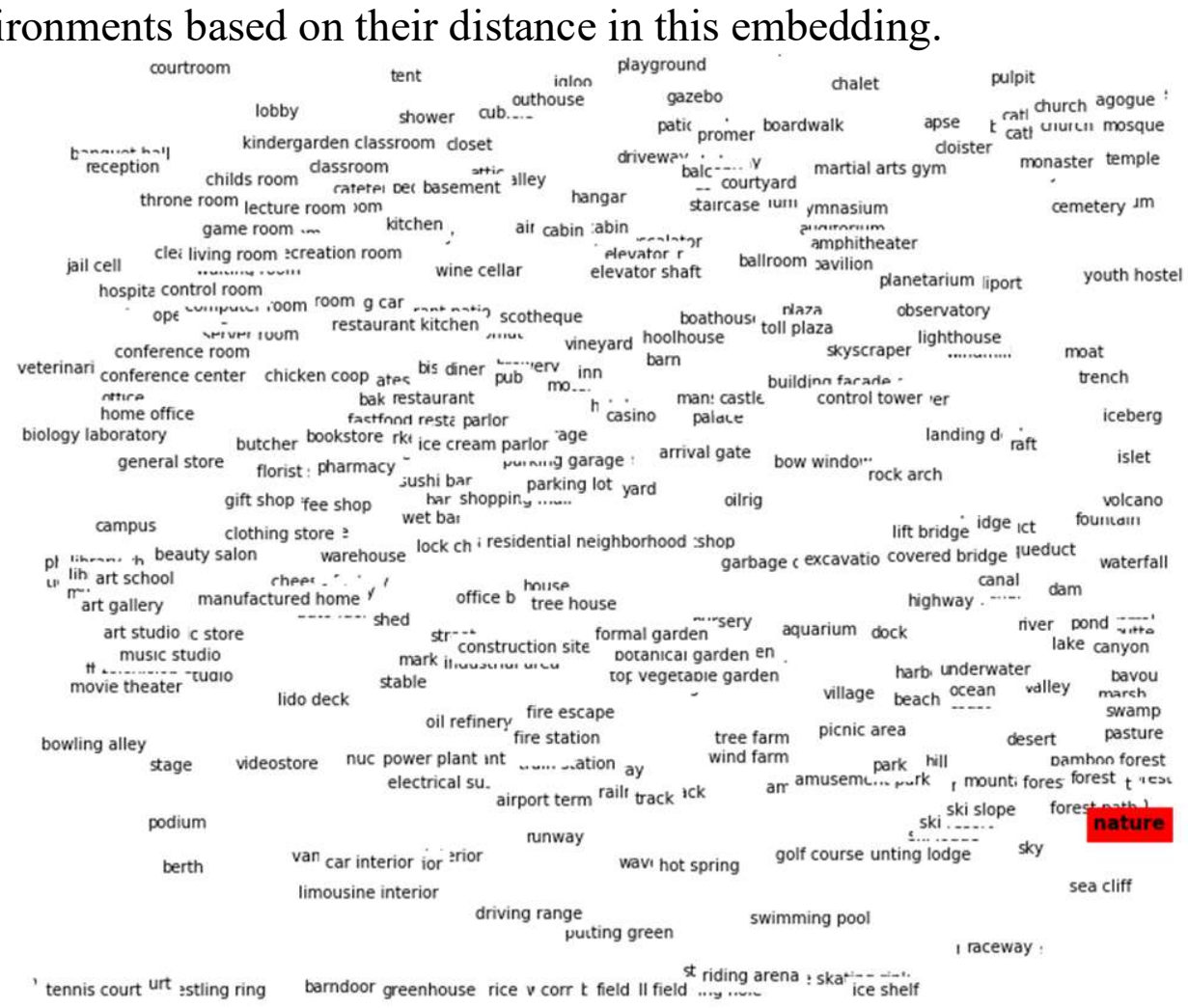

**Figure 2: Semantic relations. The environment 'nature' (red) relates to forest, mountain, sky, etc. (lower right) and not to campus, podium, car, etc. (middle).**

The semantic embedding is combined with the visual embedding (Section 3). For this combination, a normalization is needed such that metrics align, i.e., distances in each embedding should imply approximately the same difference.

## 4.2 Normalization

Normalization of both embeddings is simply done by mapping the average distance to unity. After that, both embeddings can be combined to get a 'semantic-visual embedding', illustrated in Figure 3. The actual computations to arrive at P(competent) are detailed in subsection 4.3.

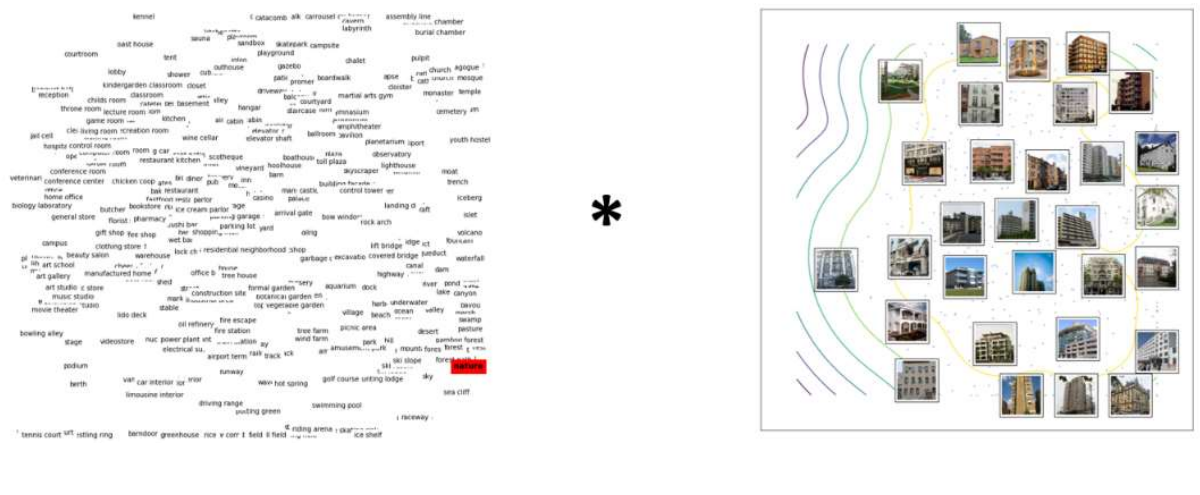

**Figure 3: Semantic and visual spaces are combined after they have been normalized to have similar metrics.**

### 4.3 P(competent)

To explain the underlying computations, we consider the example where the current environment is a computer desk, and CA-expert knows that the AI model is ‘not competent in nature’ environments. CA-expert has knowledge about many types of environments. Two examples are ‘forest’ and ‘office’. The current environment (the computer desk) is visually similar to office (high score). Semantically, the relation to ‘nature’ (in which CA-expert knows that it is not competent), is very distant (very low score). As a resultant, P(incompetent) ~ 0. The current environment looks visually not like forest (very low score), yet forest is semantically very related (high score) to nature (in which CA-expert knows it is not competent). As a resultant, P(incompetent) ~ 0. CA-expert will conclude that it is not incompetent.

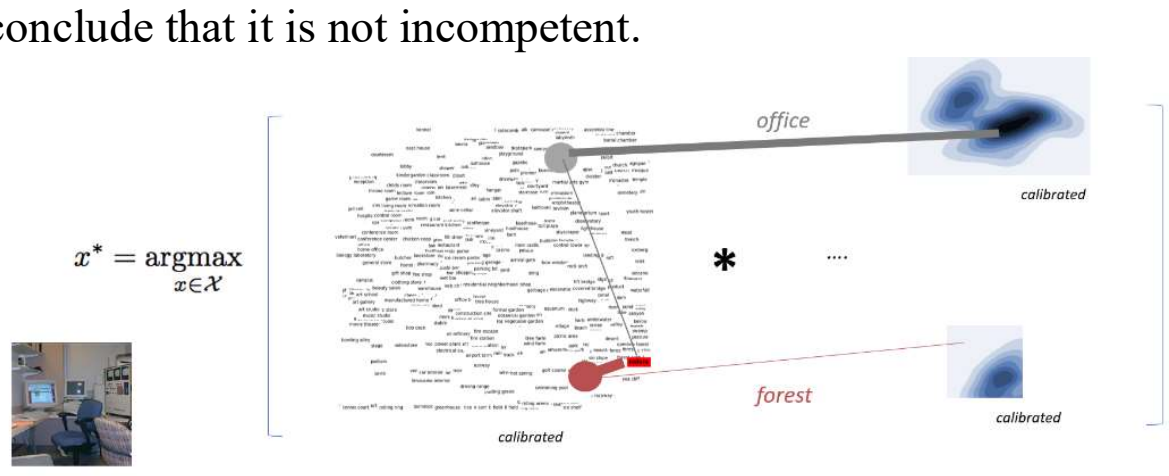

**Figure 4: Reasoning about semantic and visual relations to arrive at P(competent).**

Actually, there are much more environments in the MIT Places database than just forest and office. For all environments, this same computation is performed. That is why Figure 4 mentions ‘argmax’ over all environments. But for none it can conclude that it is incompetent.

If the next environment is a park, then the same computations are performed, but with a different outcome for P(incompetent). For park, the office environment is very unrelated, both visually and semantically (both low scores). The forest environment is related, because visually there is a reasonable relation (reasonable score), and semantically there is a strong relation (high score). Because CA-expert knows that it is ‘incompetent in nature environments’, P(incompetent) will be reasonable, according to the computations ~ 0.5 (see Figure 5).

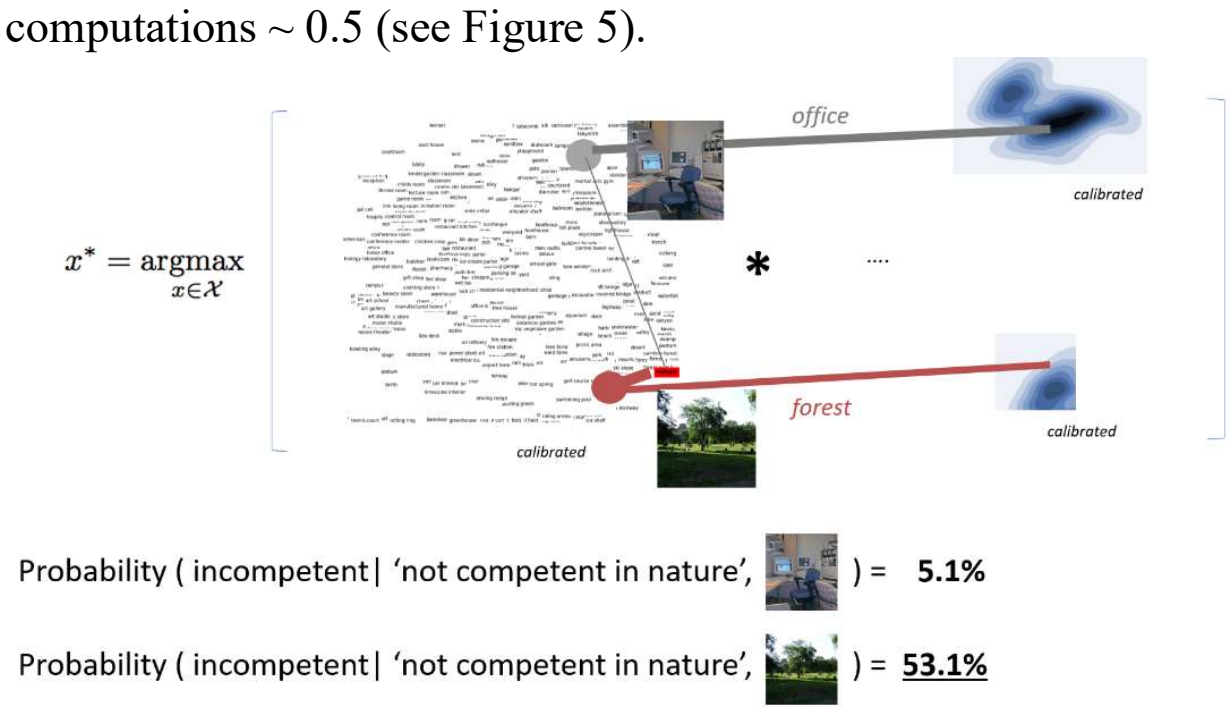

**Figure 5: For two different cases, CA-expert reasons about semantic and visual relations to arrive at P(incompetent).**

In Figure 5, both environments ‘computer desk’ and ‘park’ have different visual-semantic relations that maximize the computation rule, arriving at resp. a low P(incompetent) and high P(incompetent) for the case where the AI model is ‘incompetent in nature environments’. To conclude, CA-expert can translate high-level statements about (in)competence into computable quantities.

## 5 Results

For experiments, the Fukuoka robot dataset [9] is considered. The dataset consists of camera images recorded by a robot driving through an office building where it encounters various environments.

### 5.1 Use-case

The robot moves through corridors first, after which it enters a lab room. The goal of the robot is to find people. In the next subsections, we will show that it learns that it is competent to detect people well in corridors, where people stand upright and are clearly visible. It then learns that it is incompetent in lab rooms, because people are not well visible, occluded by chairs and desks, and in deviating poses such as sitting and hanging. The robot then performs the same route again and has updated its model of competence. It does not need human feedback about its competence. The robot now knows that it is incompetent in lab rooms, where it may solve this by asking the human for assistance.

### 5.2 Initialization and calibration

The MIT Places dataset is probed to calibrate environments, see Figure 6 for the embedding such that it knows how similar environments are and deriving the kernel size for the Gaussian pdf (see section 3.2).

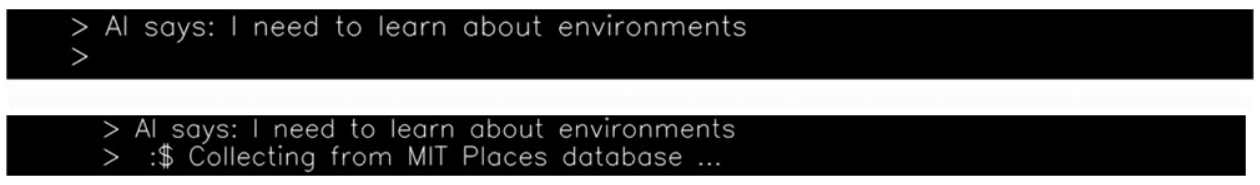

**Figure 6-A: Initialization.**

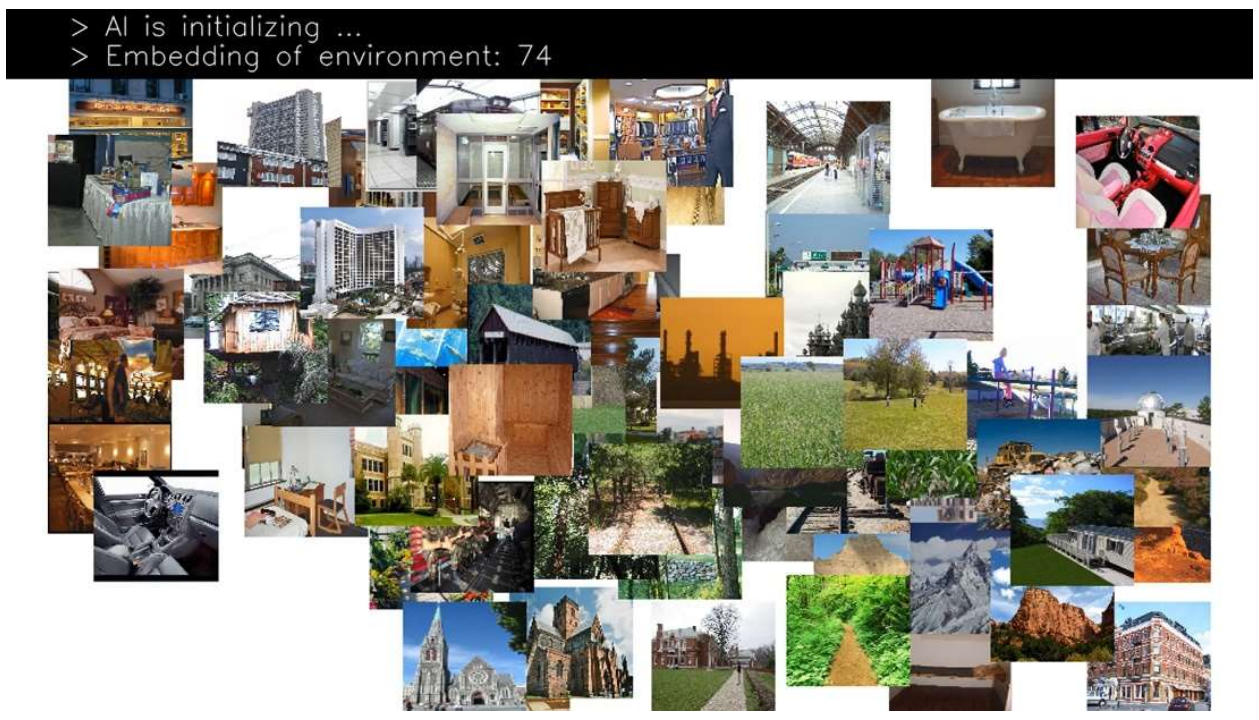

**Figure 6-B: Calibration by probing MIT Places.**

### 5.3 Unknown environment – human feedback

The robot starts driving, but it does not know anything (known ~ undefined), so it also does not know about its competence there. The human is asked for feedback about its competence, see Figure 7.

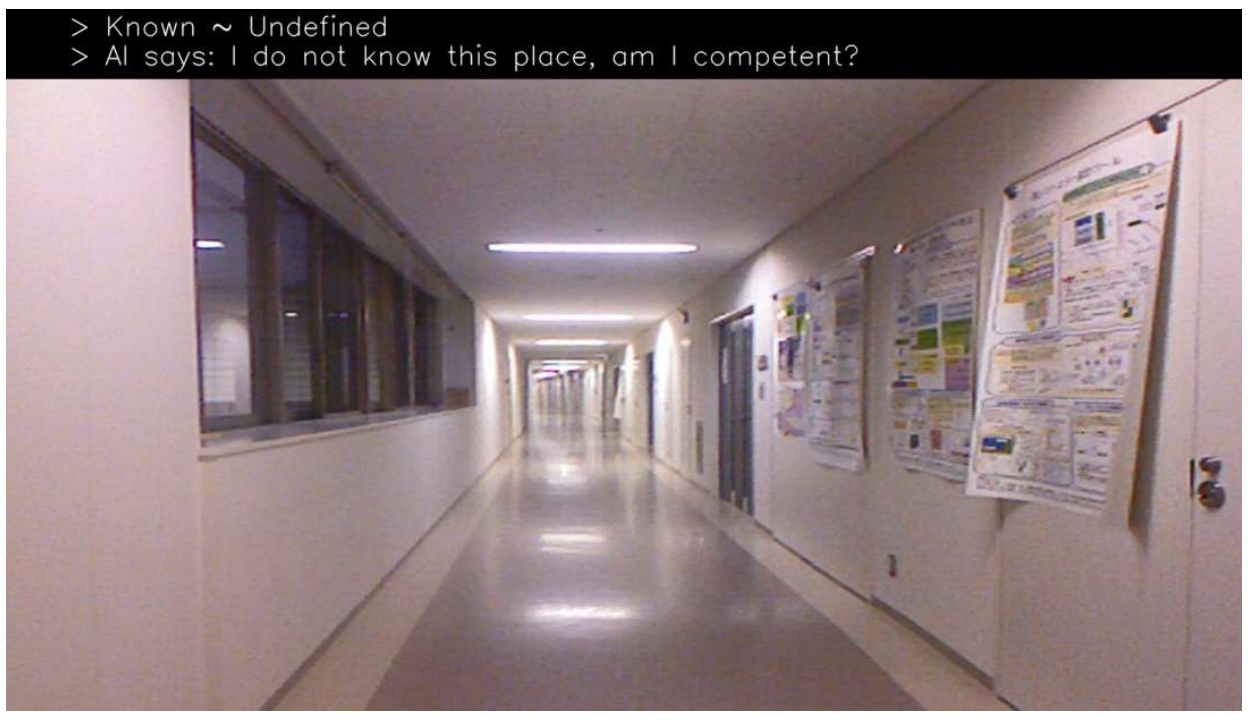

**Figure 7: Human is asked for feedback about the robot's competence.**

The human confirms that it is competent, because the robot is well able to detect people in the corridor, see Figure 8.

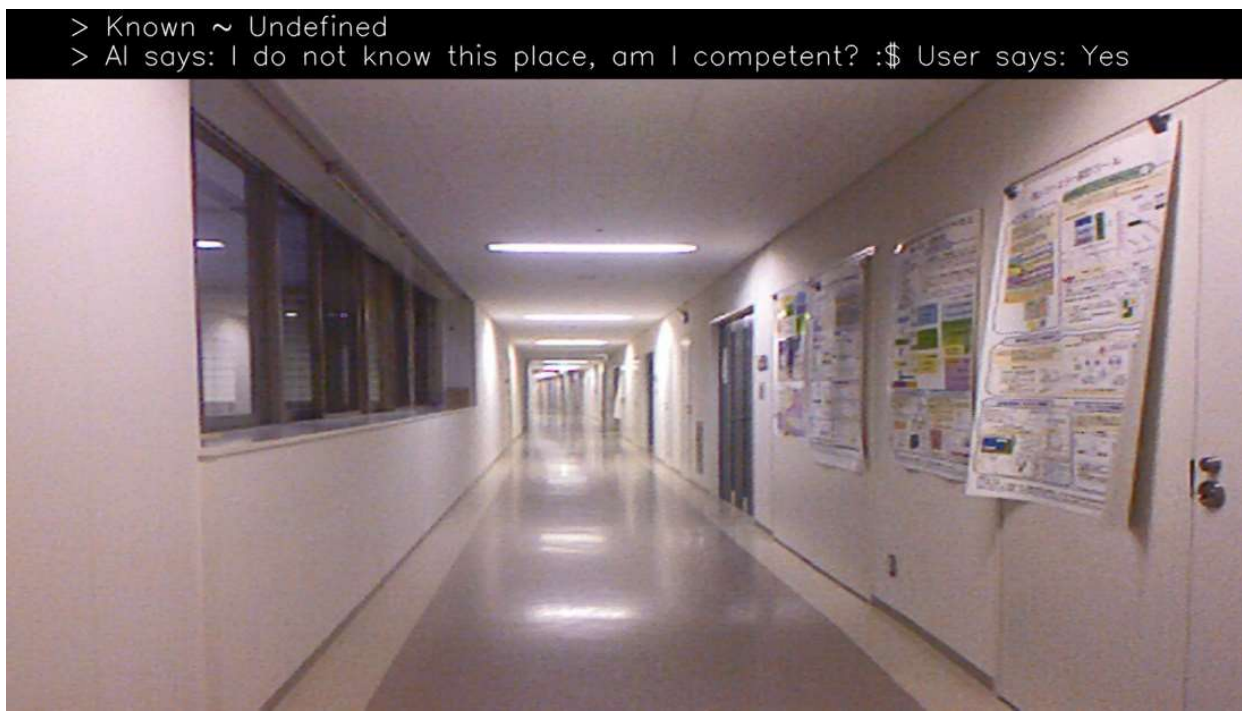

**Figure 8: Human says that the robot is competent.**

The robot knows about the remainder of the corridors, so it can generalize across similar environments without asking a human for feedback, see Figure 9.

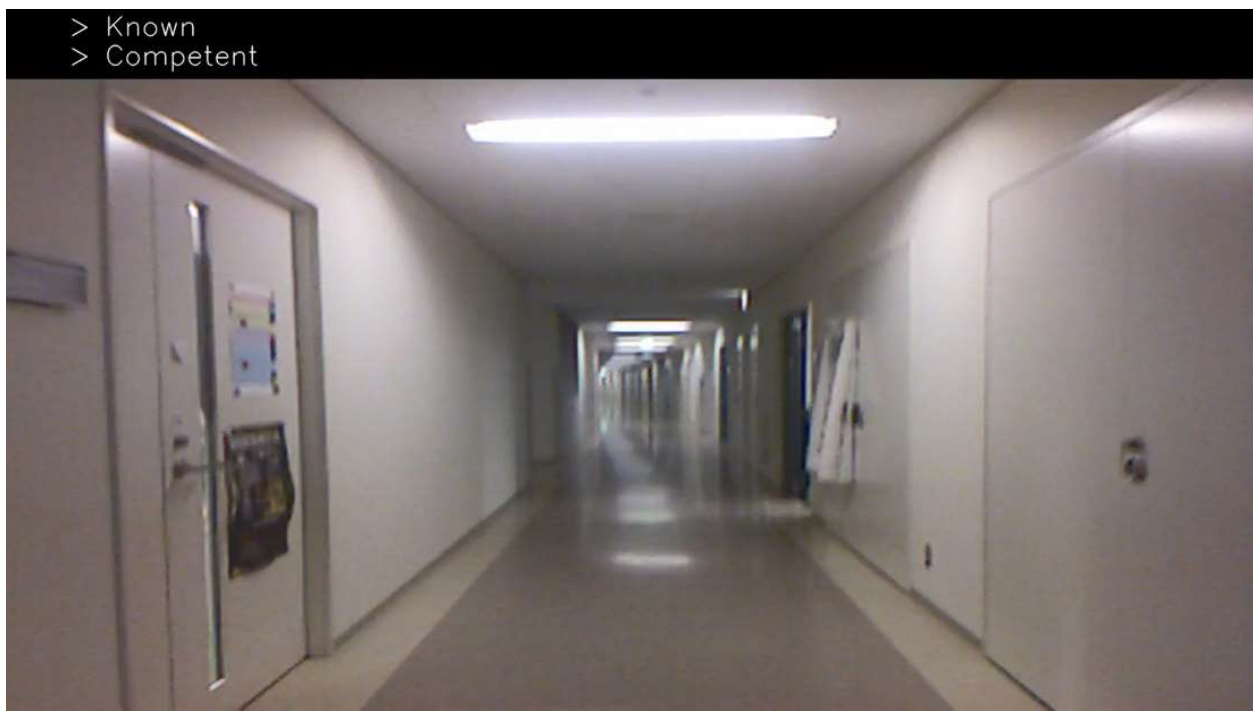

**Figure 9: Robot knows how to generalize to other corridors.**

The robot now enters a lab room, which it does not yet know, so it cannot assess its competence. Again, it asks for a human for feedback about its competence, see Figure 10.

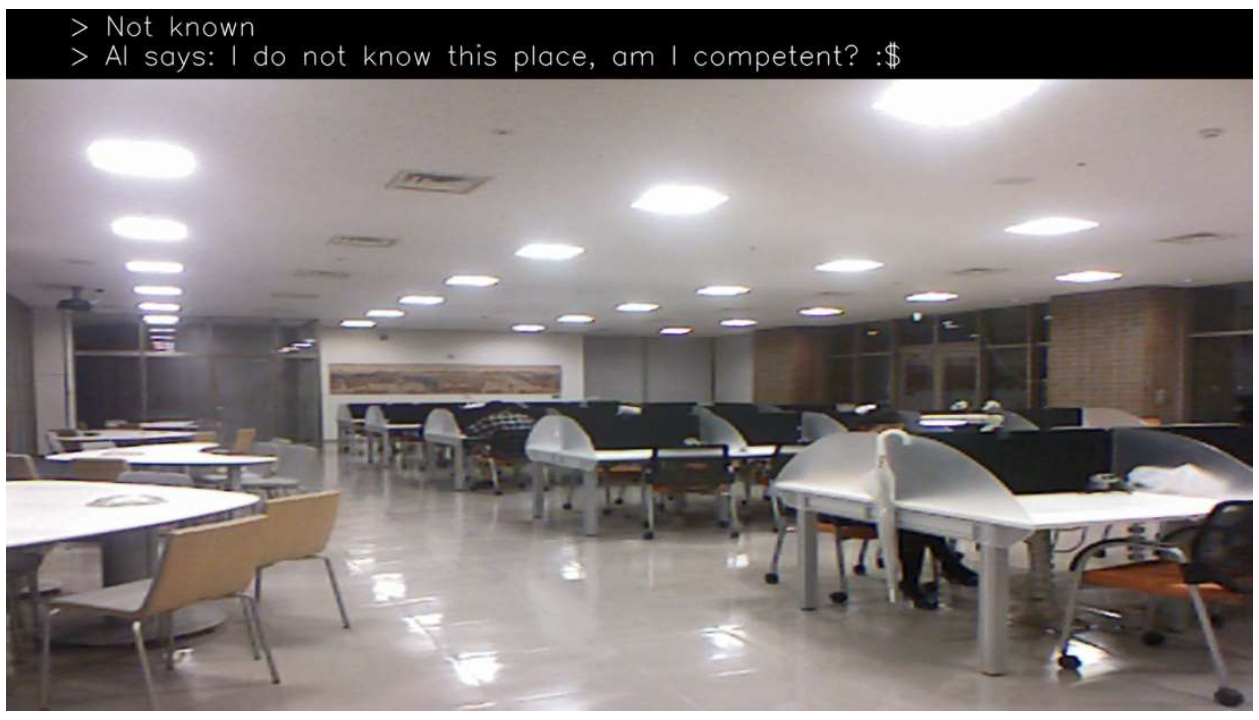

**Figure 10: Robot knows that it does not know this environment and asks the human for feedback.**

The human indicates that the robot is not competent, because it cannot detect the people in the lab room, see Figure 11.

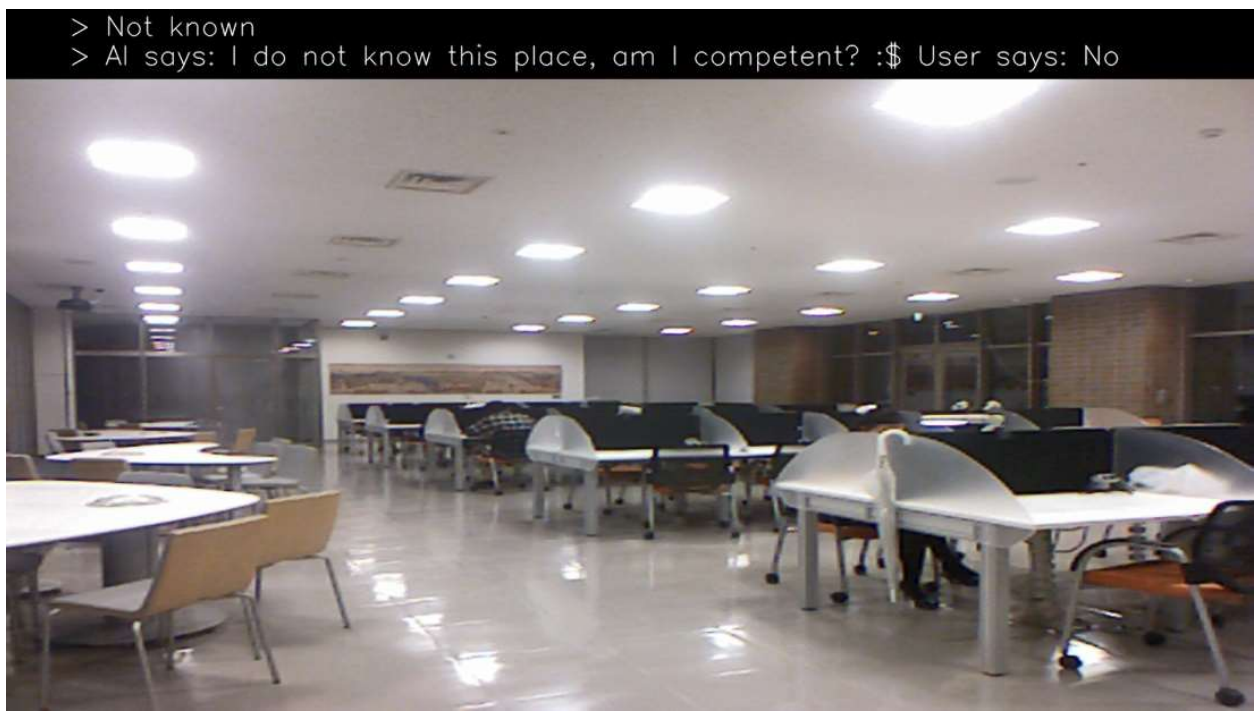

**Figure 11: Human says that the robot is not competent.**

## 5.4 Known environment – robot has updated its model of (in)competence

The robot moves again through the corridors and lab rooms. It has updated its model of competence. Hence it now knows it's (in)competence and does not need feedback from the human, see Figure 12 (corridors) and Figure 13 (lab room).

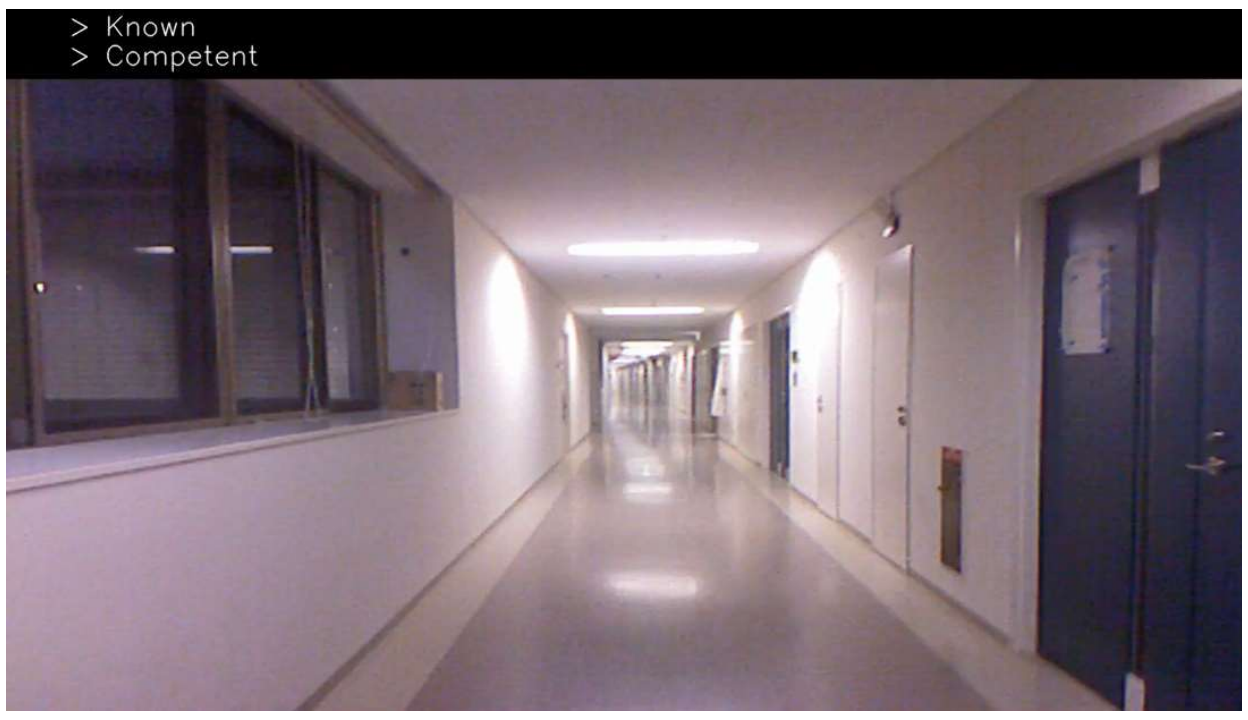

**Figure 12: Robot knows that it is competent in corridors.**

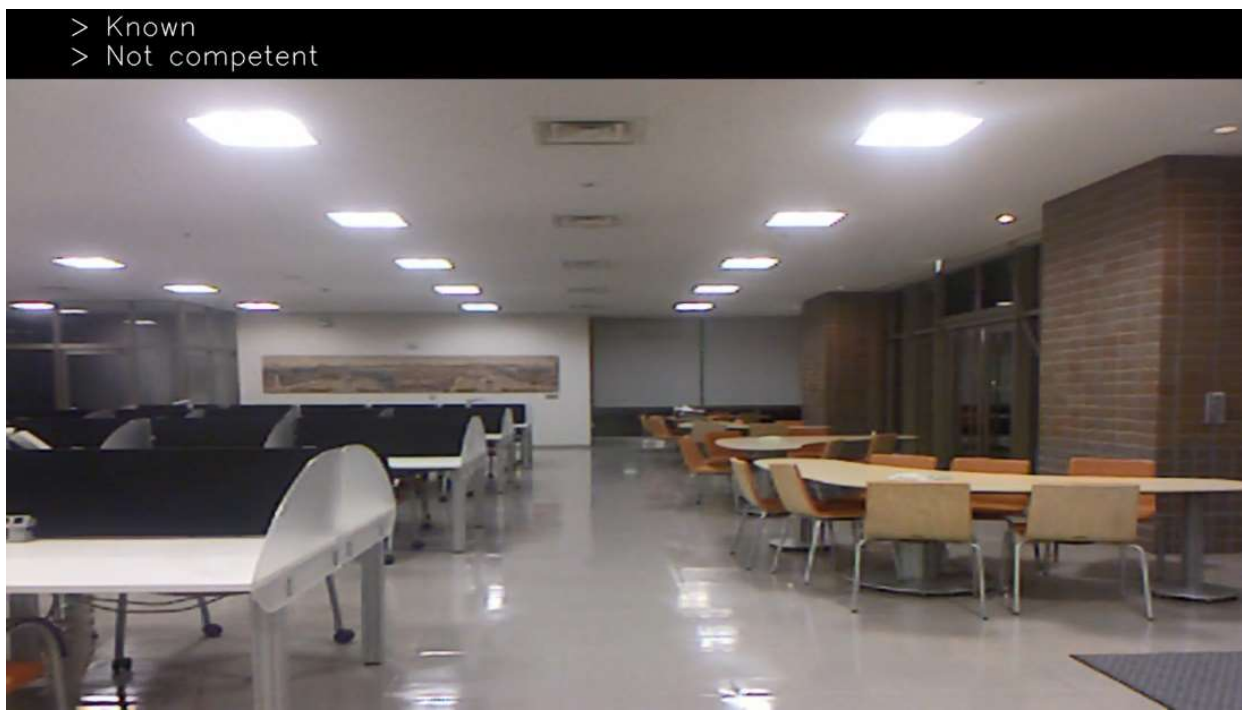

**Figure 13: Robot knows that it is not competent in lab rooms.**

## 6 Human-Robot Interaction

The experiment of the previous section is a first step towards a pervasive human-robot co-learning system, in which the human complements robot's knowledge. The next step is to implement the system in a meaningful scenario and test the learning process in a user study. This can be a robot-assisted search of "wandering visitors" in a shop or exposition center near or after closing time. Over time, the robot needs less human assistance to do its job well.

In our experiment, the robot had the initiative; human's contribution to the competence assessment and the corresponding human-robot interaction were worked out minimally. To advance human's contribution and enable a reciprocal human-agent exchange of information on robot's competences in scenarios as sketched above, we distinguish four challenges.

*First*, a more elaborate competence model (e.g., as an ontology or graph network) is needed, which is interpretable for both the human and robot. The challenge is to integrate current models into the competence assessment framework (e.g., on shared situation awareness [10]), and to present robot's environment-dependent competences to the human in an effective, usable way (e.g., integrated in operator's displays [11]). *Second*, the competence assessment must deal with uncertainties. Here, the challenge is to derive a confidence measure that is interpretable for the human, so that he or she can respond adequately [12]. *Third*, the competences of the robot will change over time and the human should adapt his or her behaviors (e.g. the feedbacks) correspondingly. The challenge is to provide insight into competence changes or dependencies that the human may not oversee or anticipate, and to establish the right trust calibration over time [13]. *Fourth*, the human should be allowed to ask "why" the robot is (in)capable to perform in a specific environment. The challenge is to develop explanations that a robot can provide, which the human needs, understands and uptakes [14,15].

## 7 Conclusions

In this paper, we have shown the importance of competence assessment for (semi-) autonomous robots, because typically the on-board AI models are not aware of their competence. We have proposed two methods for the online assessment of the competence of the AI model, respectively for the situations when nothing is known about competence beforehand, and when there is prior knowledge about competence (in semantic form). The proposed method assesses whether the current environment is known. If not, it asks a human for feedback about its competence. If it knows the environment, it assesses its competence by generalizing from earlier experience. Results on real data show the merit of competence assessment for a robot moving through corridors, where it is competent, and through lab rooms, where it is not competent. We have shown how the robot interacts with a human to ask for feedback about competence in various environments. We have demonstrated how the robot is able to generalize (e.g., to other corridors), in order to limit human involvement, and that it is able to assess its (in)competence after the human feedback and update its model of competence accordingly.

We have discussed the role of the human in the competence assessment of a robot, distinguishing four research challenges. By meeting these challenges, we aim to establish a form of human-AI co-learning that harmonizes human and artificial intelligence progressively [16]. For the future, we aim at co-learning that entails competence models of both the robot and the human [17].

## REFERENCES

[1] Mei Wang, Weihong Deng, Deep Visual Domain Adaptation: A Survey, Neurocomputing 2018.
[2] Xingchao Peng, Qinxun Bai, Xide Xia, Zijun Huang, Kate Saenko, Bo Wang, Moment Matching for Multi-Source Domain Adaptation, ICCV 2019.

[3] Kimin Lee, Kibok Lee, Honglak Lee, Jinwoo Shin, A Simple Unified Framework for DetectingOut-of-Distribution Samples and Adversarial Attacks, NeurIPS 2018.

[4] Guo, C., Pleiss, G., Sun, Y., & Weinberger, K. Q., On Calibration of Modern Neural Networks, ICML 2017.

[5] Mark Sandler, Andrew Howard, Menglong Zhu, Andrey Zhmoginov, Liang-Chieh Chen, MobileNetV2: Inverted Residuals and Linear Bottlenecks, CVPR 2018.

[6] Christopher M. Bishop, Pattern Recognition and Machine Learning, Springer-Verlag, 2006.

[7] B. Zhou, A. Lapedriza, A. Khosla, A. Oliva, and A. Torralba, Places: A 10 million Image Database for Scene Recognition, PAMI 2017.

[8] Mikolov, Tomas; Sutskever, Ilya; Chen, Kai; Corrado, Greg S.; Dean, Jeff, Distributed Representations of Words and Phrases and their Compositionality, NeurIPS 2013.

[9] http://robotics.ait.kyushu-u.ac.jp/kyushu_datasets/indoor_rgbd.html

[10] Smets, N. J., Neerincx, M. A., Jonker, C. M., & Båberg, F. (2017). Ontology-based situation awareness support for shared control. In *Proceedings of the Companion of the 2017 ACM/IEEE International Conference on Human-Robot Interaction* (pp. 289-290).

[11] Parasuraman, R., Caccamo, S., Båberg, F., Ögren, P., & Neerincx, M. (2017). A new UGV teleoperation interface for improved awareness of network connectivity and physical surroundings. *Journal of Human-Robot Interaction*, *6*(3), 48-70.

[12} van der Waa, J., van Diggelen, J., & Neerincx, M. (2018). The design and validation of an intuitive confidence measure. *memory*, *2*, 1.

[13] de Visser, E. J., Peeters, M. M., Jung, M. F., Kohn, S., Shaw, T. H., Pak, R., & Neerincx, M. A. (2019). Towards a Theory of Longitudinal Trust Calibration in Human–Robot Teams. *International Journal of Social Robotics*, 1-20.

[14] Miller, T. (2019). Explanation in artificial intelligence: Insights from the social sciences. *Artificial Intelligence*, *267*, 1-38.

[15] Neerincx, M. A., van der Waa, J., Kaptein, F., & van Diggelen, J. (2018). Using perceptual and cognitive explanations for enhanced human-agent team performance. In *International Conference on Engineering Psychology and Cognitive Ergonomics* (pp. 204-214). Springer, Cham.

[16] van den Bosch, K., Schoonderwoerd, T., Blankendaal, R., & Neerincx, M. (2019). Six Challenges for Human-AI Co-learning. In *International Conference on Human-Computer Interaction* (pp. 572-589). Springer, Cham.

[17] Mioch, T., Kroon, L., & Neerincx, M. A. (2017). Driver readiness model for regulating the transfer from automation to human control. In *Proceedings of the 22nd international conference on intelligent user interfaces* (pp. 205-213).